\documentclass[10pt,twocolumn,letterpaper]{article}

\usepackage[pagenumbers]{cvpr} 

\usepackage{graphicx}
\usepackage{amsmath}
\usepackage{amssymb}
\usepackage{bm}
\usepackage{booktabs}
\usepackage{comment}
\usepackage{xcolor}
\usepackage{siunitx}
\graphicspath{{FiguresRainbow/}}
\def\capfont{\normalfont\small}
\usepackage[percent]{overpic}
\usepackage{adjustbox}

\newif\ifcolors
\colorsfalse

\ifcolors
    \newcommand{\red}[1]{{\color{red}#1}}
    
    \newcommand{\MS}[1]{\textcolor{orange}{#1}}

\else
    \newcommand{\red}[1]{#1}
    \newcommand{\MS}[1]{#1}

\fi

\newcommand{\figTeaser}{
\begin{figure}[t]
    \centering
    \includegraphics[width=\columnwidth]{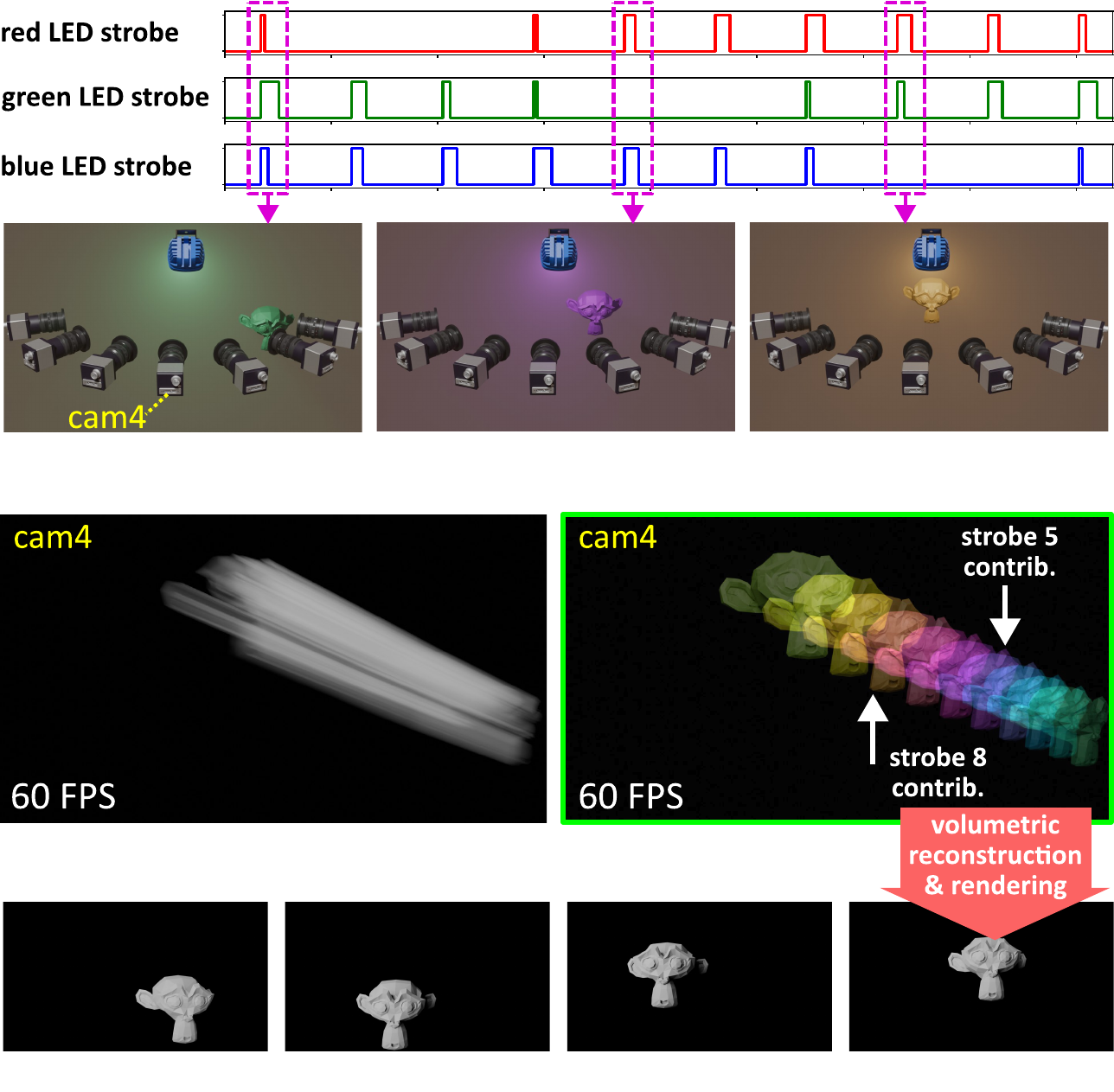}
    \put(-230,132){\capfont \textbf{(a)} high-speed color strobing during a single camera exposure}
    \put(-224,49){\capfont \textbf{(b)} constant light frame}
    \put(-112,49){\capfont \textbf{(c)} strobed frames}
    \put(-185,0){\capfont \textbf{(d)} novel views rendered at 600 FPS}
    \caption{High-Speed volumetric scene encoding and reconstruction. \textbf{(a)} We strobe a scene having rapid motion with a sequence of 10 distinct colors during each captured frame. \textbf{(b)} With constant illumination, the object motion would result in a motion-blurred frame, where the exact object motion is obscured.
    \textbf{(c)} With our system, however, the resulting frames per camera contain a colorful mixture of the object's intermediate frames.
    \textbf{(d)} The individual strobed frames from all cameras are then used to recover a volumetric dynamic representation of the scene, enabling novel view synthesis at a high frame rate of 600 Hz.} 
    \label{fig:opening}
\end{figure}
}

\newcommand{\figStrob}{
\begin{figure}[t]
  \centering
  \begin{overpic}[width=1.0\linewidth]{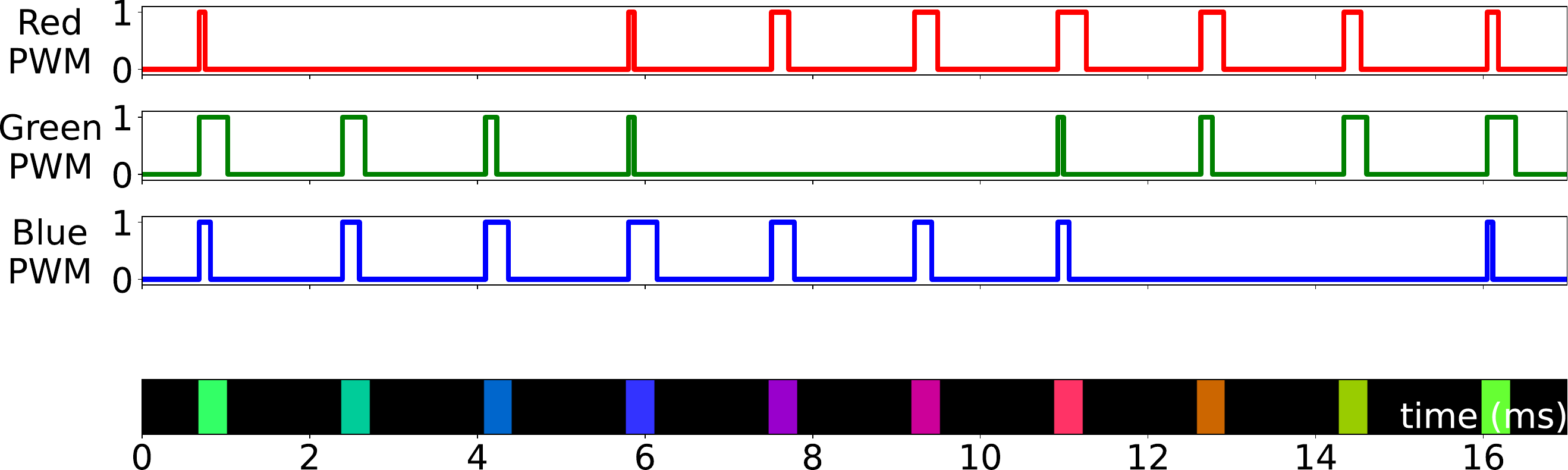}
    \put(53.5, 9){\makebox(0,0)[c]{\capfont \textbf{(a)} strobing scheme }}
    \put(53.5, -3){\makebox(0,0)[c]{\capfont \textbf{(b)} resulting color}}
  \end{overpic}
  \vspace{1pt}
  \caption{Generating colors by pulse modulation. We illuminate the scene with three direct-color LEDs. \textbf{(a)} Each LED is driven by an Arduino using on/off digital pulses whose widths set the per-strobe color. \textbf{(b)} Because each strobe is short compared to scene motion, the object appears static during a strobe, and the integrated LED pulses produce the desired color.
  }
  \vspace*{-4mm}
  \label{fig:pwm_to_color}
\end{figure}
}

\newcommand{\figChopper}{
\begin{figure*}[t]
  \centering
  \begin{overpic}[width=0.8\linewidth]{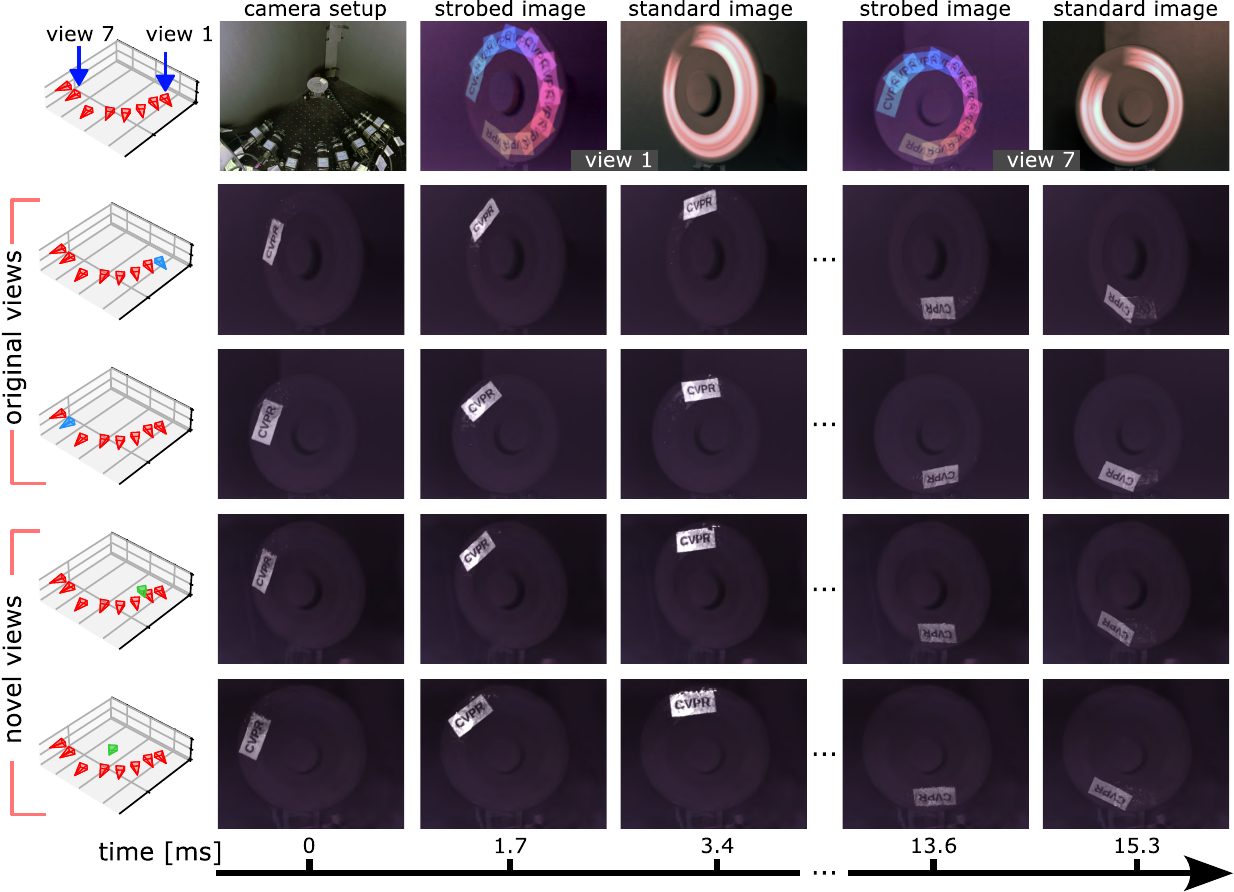}
  \end{overpic}
  \caption{Experimental results on spinning disk. We capture a rapidly spinning disk using eight cameras. \textbf{(Top row)} shows the experiment setup and the raw input frames captured from two of the cameras' views. On the right of each strobed input image is a same-exposure image without strobing, showing the resulting images without color encoding.
  \textbf{(Middle rows)} Several high-speed frames from the original views. Note that our method correctly decodes the high-speed frames from the color-encoded frames, yielding high-speed multi-view videos.
  \textbf{(Bottom rows)} Several high-speed frames of the spinning disk from \textit{novel views} rendering using our modified Gaussian-Flow.
  }  
  \label{fig:chooper}
\end{figure*}
}

\newcommand{\figNerf}{
\begin{figure*}[t]
  \centering
  \begin{overpic}[width=0.8\linewidth]{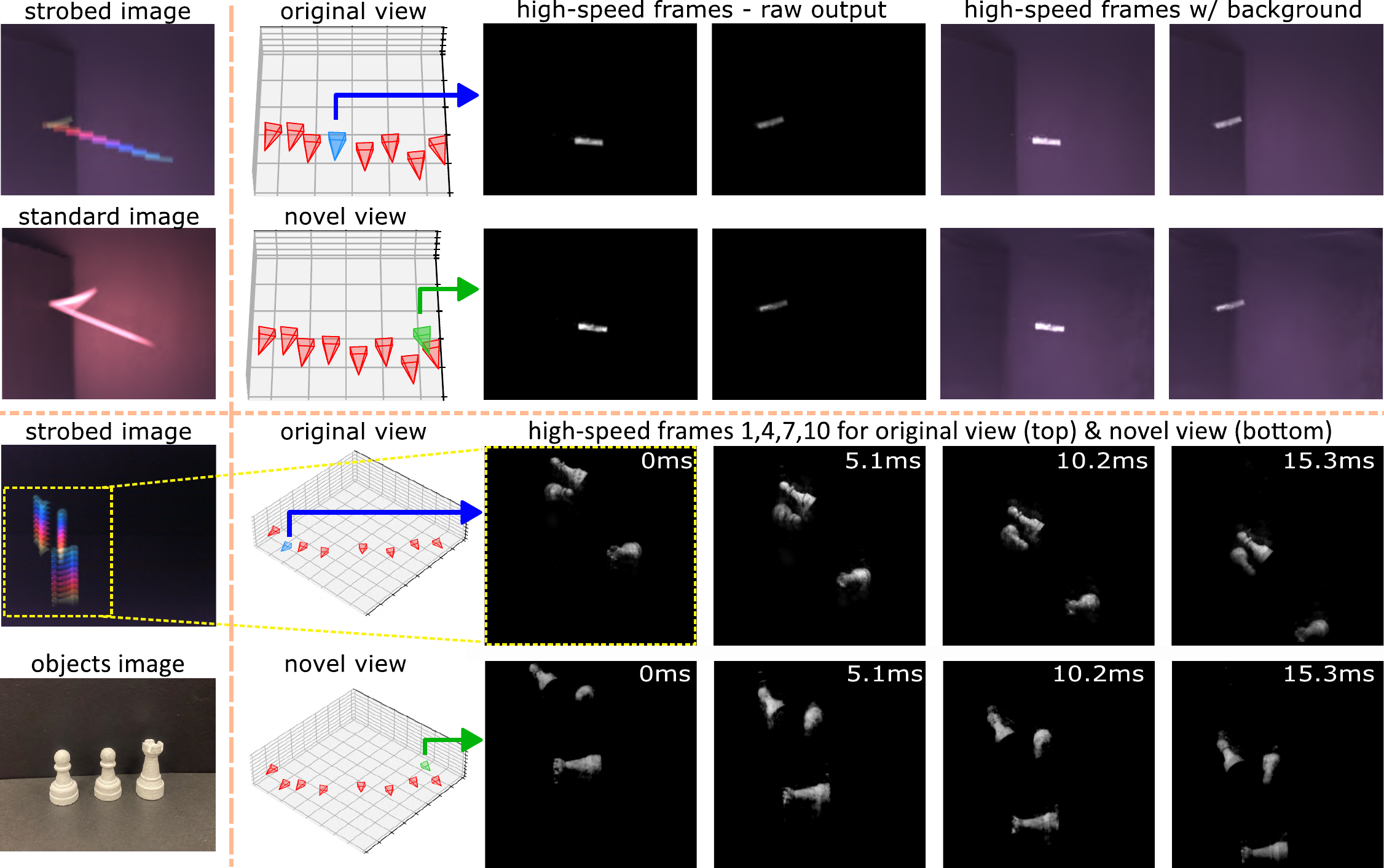}
  \end{overpic}
  \caption{Motion capture experiments. \textbf{(Top)} Nerf dart experiment showing the strobed and non-strobed frames as well as recovered high-speed interframes.
  \textbf{(Bottom)} Flying chess pieces experiment; The rows show the recovered interframes for an existing and novel view.
  }  
  \label{fig:nerf}
  \vspace{-15pt}
\end{figure*}
}

\newcommand{\figSimulation}{
\begin{figure}[t]
    \centering
    \includegraphics[width=\columnwidth]{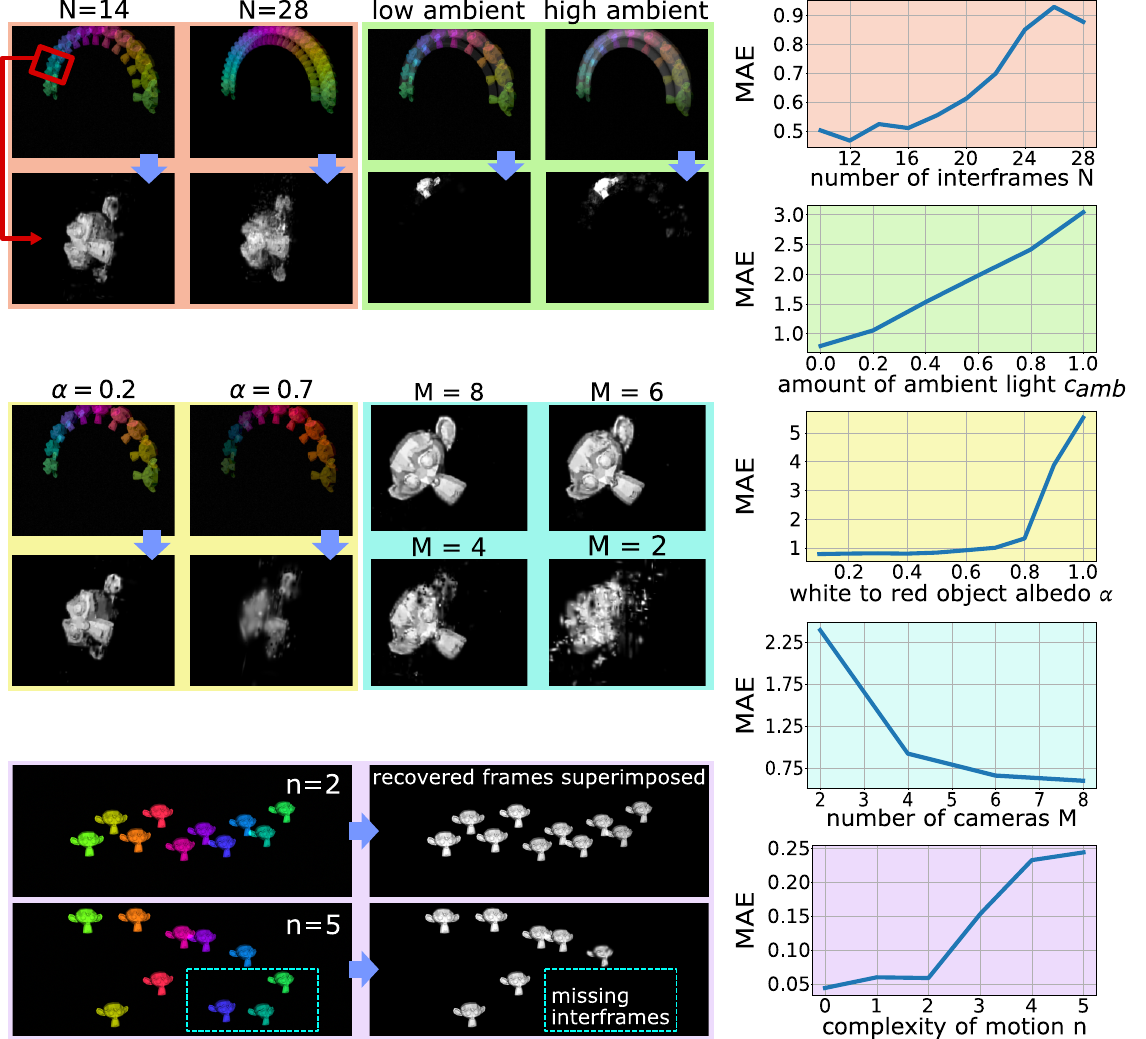}
     \put(-204.5,142){\capfont\textbf{(a)}}
  \put(-131,142){\capfont\textbf{(b)}}
  \put(-204.5,63.5){\capfont\textbf{(c)}}
  \put(-131,63.5){\capfont\textbf{(d)}}
  \put(-167.5,-8){\capfont\textbf{(e)}}
  \put(-9,197.75){\capfont\textbf{(a)}}
  \put(-9,155){\capfont\textbf{(b)}}
  \put(-9,112.5){\capfont\textbf{(c)}}
  \put(-9,67){\capfont\textbf{(d)}}
  \put(-9,22.5){\capfont\textbf{(e)}}
    \caption{Simulation performance analysis. The left columns show example recovered frames; the right column shows MAE plots.
    \textbf{(a)} Effect of interframe number $N$.
    \textbf{(b)} Effect of ambient-light level (x-axis: ambient light as a percentage of total signal).
    \textbf{(c)} Effect of non-white albedo (x-axis: blend ratio between pure-white and pure-red reflectance).
    \textbf{(d)} Reconstruction quality versus the number of cameras.
    \red{ \textbf{(e)} Effect of the motion complexity.}
    }
    \vspace{-15pt}
    \label{fig:simulation}
\end{figure}
}

\newcommand{\figHW}{
\begin{figure}[t]
    \centering
    \includegraphics[width=0.7\columnwidth]{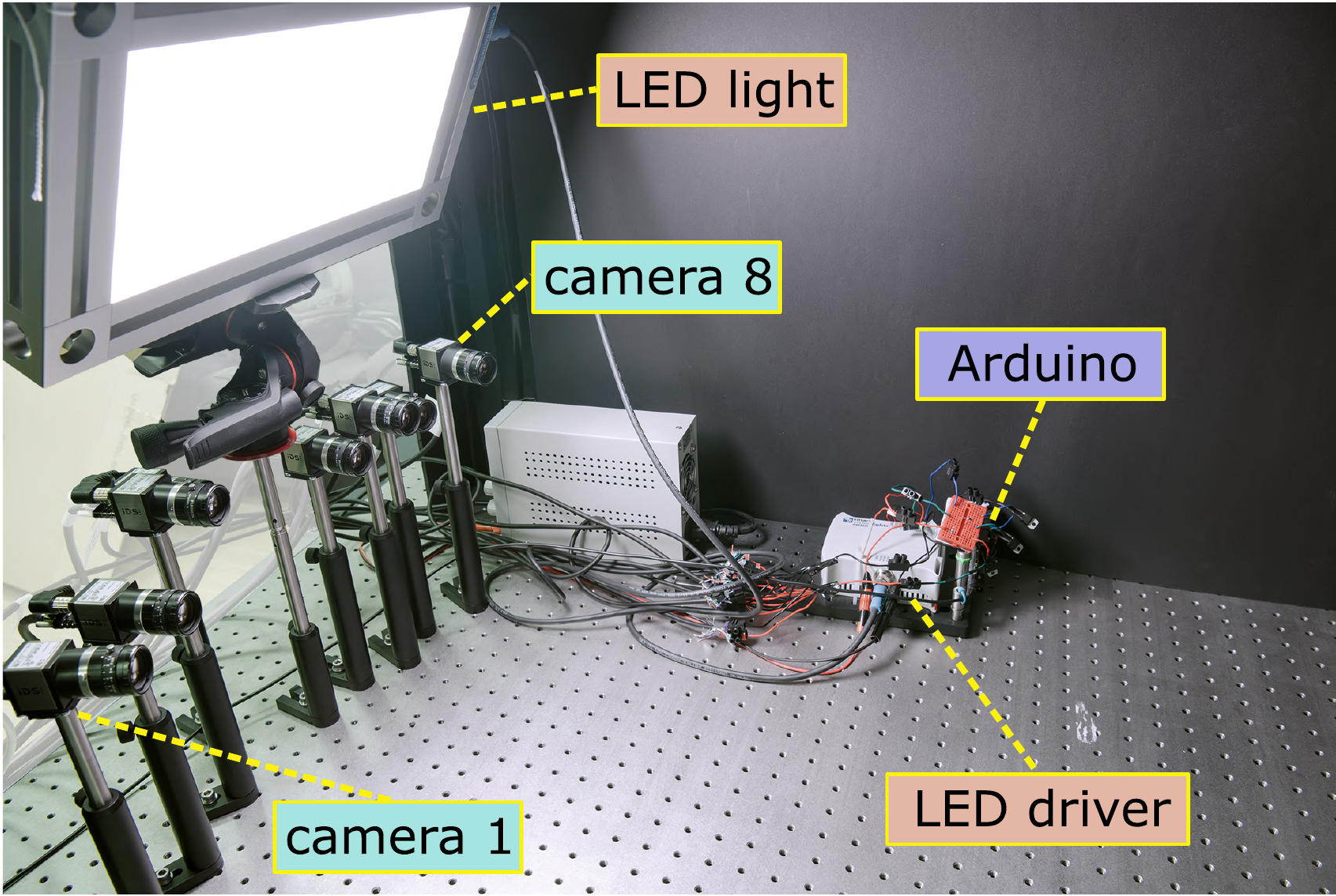}
    \caption{Experimental prototype. Our prototype uses three LED channels (R/G/B) driven by an LED driver and controlled via Arduino PWM. Each channel is strobed independently at high speed.}
    \vspace{-15pt}
    \label{fig:HW}
\end{figure}
}

\newcommand{\ledR}{{\bm c}_{\rm R}} 
\newcommand{\ledG}{{\bm c}_{\rm G}} 
\newcommand{\ledB}{{\bm c}_{\rm B}} 
\newcommand{\ledRGB}{{\bm c}_{\rm RGB}} 
\newcommand{\ledRGBun}{\tilde{\bm c}_{\rm RGB}} 

\newcommand{\pixel}{{\bf x}} 

\newcommand{\imgc}{I^{\rm rgb}(\pixel,c)} 
\newcommand{\imgeilon}{I^{\rm rgb}}
\newcommand{\imgeilone}{\hat{I}^{\rm rgb}}
\newcommand{\imgi}{I^{\rm int}(\pixel,t)} 
\newcommand{\imgn}{I^{\rm int}(\pixel,t_n)}

\newcommand{\amat}{\mathbf{c}_{\textnormal{RGB}}^n(c)}

\definecolor{cvprblue}{rgb}{0.21,0.49,0.74}
\usepackage[pagebackref,breaklinks,colorlinks,allcolors=cvprblue]{hyperref}

\def\paperID{6578} 
\def\confName{CVPR}
\def\confYear{2026}

\title{Color-Encoded Illumination for High-Speed Volumetric Scene Reconstruction}
\author{David Novikov \qquad Eilon Vaknin \qquad Narek Tumanyan\qquad Mark Sheinin\\
Weizmann Institute of Science, Israel\\
{\tt\small \{david.novikov, eilon.vaknin, narek.tumanyan, mark.sheinin\}@weizmann.ac.il}\\
{\small
\url{https://davidnovikov.github.io/color-encoded-illumination-website/}}
}

\begin{document}
\maketitle

\begin{abstract}
The task of capturing and rendering 3D dynamic scenes from 2D images has become increasingly popular in recent years.
However, most conventional cameras are bandwidth-limited to 30–60 FPS, restricting these methods to static or slowly evolving scenes.
While overcoming bandwidth limitations is difficult for general scenes, recent years have seen a flurry of computational imaging methods that yield high-speed videos using conventional cameras for specific applications (e.g., motion capture and particle image velocimetry).
However, most of these methods require modifications to a camera's optics or the addition of mechanically moving components, limiting them to a single-view high-speed capture. Consequently, these methods cannot be readily used to capture a 3D representation of rapid scene motion. In this paper, we propose a novel method to capture and reconstruct a volumetric representation of a high-speed scene using only unaugmented low-speed cameras. Instead of modifying the hardware or optics of each individual camera, we encode high-speed scene dynamics by illuminating the scene with a rapid, sequential
color \red{coded} sequence. 
This results in simultaneous multi-view capture of the scene, where high-speed temporal information is encoded in the spatial intensity and color variations of the captured images.
To construct a high-speed volumetric representation of the dynamic scene, we develop a novel dynamic Gaussian Splatting-based approach that decodes the temporal information from the images. 
We evaluate our approach on simulated scenes and real-world experiments using a multi-camera imaging setup, showing first-of-a-kind high-speed volumetric scene reconstructions.

\end{abstract}    

\section{Introduction}
\label{sec:intro}
\figTeaser

The capture speed of any camera is limited by its data bandwidth, namely the rate at which its readout electronics can transfer image data from the sensor to its memory. For general scenes, the inherent bandwidth limitation is hard to overcome without resorting to expensive, specialized high-speed cameras. Moreover, most high-speed cameras suffer from a poor `temporal dynamic range', namely that the number of \textit{total} frames that can be captured at high frame rates is low (\eg, 256), severely limiting the capture duration and requiring precise triggering to capture the desired event \cite{youtube:7i8E6uQQtog}. However, some special yet important imaging scenarios like motion capture, particle imaging velocimetry (PIV), ballistics tracking, and other high-speed analytical techniques, have inherent characteristics that lend themselves to computationally encoding high-speed videos within the limited bandwidth. \red{Our method targets these scenarios in which the objects' high-speed motion, rather than their appearance, is of primary interest.}

There are two main strategies for `squeezing' higher frame rates from conventional, low-speed cameras. The first involves encoding full video frames into a subset of camera pixels, thereby increasing the frame rate by the full-frame-to-subset ratio  \cite{antipa2019video, weinberg2020100,monakhova2021untrained,tandi2025rngcam,Sheinin:2021:Deconv}. 
The second strategy is multiplexing several high-speed frames into a single low-speed frame, then computationally decomposing the captured (multiplexed) frame into a high-speed video \cite{chan2023spincam,hitomi2011video}. 
Nevertheless, most prior works, irrespective of the frame encoding process, focus on recovering a single-view high-speed video of the scene. 
Moreover, since most prior approaches require specialized hardware to achieve the desired multiplexing (\eg, spatial light modulators, novel sensors with per-pixel control, optical components such as diffraction gratings, diffusers, and more), scaling these approaches to multi-view scenes would require replicating and calibrating the complex imaging system many times over, as well as precise temporal synchronization.

In this work, we focus on recovering volumetric representations of high-speed scenes using low-speed cameras. We follow the second strategy by encoding multiple video frames into a single frame. Similarly to the work of \citet{chan2023spincam}, we use \textit{color} to encode multiple interframes\footnote{We borrow the term `interframes' from the video compression field since it is elegantly more concise than `intermediate frames'.}
during the camera's exposure period for each low-speed frame. However, unlike their method, which temporally augments the camera's point-spread function (PSF) using a spinning diffraction grating, we augment \textit{the scene's} object color by strobing it with a pre-defined color sequence (see Fig.~\ref{fig:opening}). The resulting low-speed frames contain a linear mixture of different timestamps with distinct colors. Like some prior works \citep{xiong2017rainbow,stroboscopic_microscopy_3_hue}, our method encodes the `timestamp' information \textit{directly within the scene} (via temporal color projection), making it invariant to the camera optics or type.

Our approach lends itself to volumetric high-speed capture since it can be applied to \textit{multiple} unaugmented cameras simultaneously using a single scene-strobing illuminant. To recover the high-speed scene, we develop a dynamic Gaussian Splatting-based method that takes the color-strobed low-speed camera frames as input and fits a dynamic Gaussian splatting model to the high-speed 3D scene. Our optimization decodes the high-speed volumetric scene while leveraging the multi-view geometric data to constrain the result. To the best of our knowledge, our method is the first to tie together volumetric rendering with compressive high-speed imaging. 


We evaluated our approach on simulated data and in real-world experiments using a hardware prototype, demonstrating its ability to capture rapidly moving objects. 
We present first-of-a-kind volumetric renderings of high-speed motion captured at 60 frames per second (FPS) and decoded to 600 FPS, revealing the dynamics of fast objects that cameras cannot capture at their maximum speed. We also use simulated data to conduct a performance analysis, demonstrating how different key factors in our approach affect its performance. 
Our method takes the first step toward unifying volumetric rendering with high-speed compressive video and will pave the way for future work in this domain.

\section{Related works}
\label{sec:related}
\subsection{High-speed computational imaging}
Many prior works have focused on achieving high-speed capture without using expensive, bulky, high-bandwidth cameras, instead coding the camera exposure at the aperture or pixel level \cite{raskar2006coded,holloway2012flutter,reddy2011p2c2,martel2020neural}, with pre-defined \cite{reddy2011p2c2} or random codes \cite{portz2013random}. Many works also exploited the rapid inter-row rolling shutter rate to encode scene frames into individual rows \cite{Sheinin:2021:Deconv,antipa2019video,gu2010coded,weinberg2020100}.      
Later, works began to use emerging imaging technologies to capture high-speed video or enhance traditional camera footage. Event cameras, for example, were extensively investigated to improve low-speed conventional videos captured simultaneously \cite{tulyakov2021time,tulyakov2022time}. 

A common feature of nearly all prior works is that they use specialized hardware to either augment a conventional camera (\eg, coded aperture, coded pixels, an added diffuser, or diffraction gratings), or to capture the scene itself. This limits such systems to single-camera capture, whereas simultaneous multi-camera capture would require replicating specialized, and sometimes costly, prototypes. Conversely, our method applies \textit{no} camera augmentation. It can thus trivially accommodate an arbitrary number of cameras simultaneously capturing the scene. \footnote{Our method requires color cameras with global shutter sensors.}

\subsection{Visual coding using color}
A subset of prior works used \textit{color}, in particular, to encode spatio-temporal information. 
Xiong \etal, used the light exiting a diffraction grating to color particles at different depth planes for particle image velocimetry (PIV) \cite{rainbow_piv_4}. Sheinin \etal, used the diffracted light entering a camera to encode the sparse positions of scene objects \cite{Sheinin:2021:Deconv} and later more general scenes \cite{chan2023spincam}. Tippur \etal, used color to encode the position of deformation of a tactile sensor to allow finer texture analysis of objects being grabbed \cite{tactile_sensor_6}.
Jaques \etal, used red, green, and blue illumination to upscale the temporal video resolution by a factor of three using each color channel as an interframe \cite{stroboscopic_microscopy_3_hue}.
Recently, Verma \etal, combined flashing multi-spectral LEDs with a rolling shutter sensor to capture hyperspectral images \cite{rolling_shutter_hyper_spectral}.
Chan \etal, used a rotating diffraction grating to produce a time-varying point spread function \cite{chan2023spincam}. Like them, we encode many interframes into a single exposure, but instead of modulating light in the sensor domain, we modify the appearance of scene objects, enabling multi-camera capture and, therefore, multi-view volumetric reconstruction. Lastly, Veeraraghavan \etal used color strobing for high-speed periodic scene sensing \cite{periodic_stroboscopic_7}, whereas we target general, aperiodic scenes.

\subsection{Dynamic 3D reconstruction}
Recently, many methods have tackled dynamic 3D reconstruction from single- or multi-view video.
Implicit representation-based methods \cite{DNeRFNR, hypernerf, Cao2023HEXPLANE, li2020neural, park2021nerfies, li2023dynibar} use volumetric rendering, typically with NeRF, along with an optimizable neural deformation field to reconstruct the dynamic scene. After the invention of 3D Gaussian Splatting \cite{kerbl20233d}, the trend has shifted towards using explicit representations of dynamic scenes \cite{luiten2024dynamic}. Some of the 3DGS-based methods optimize for canonical Gaussians along with a learned deformation field that determines the motion of each Gaussian \cite{yang2024deformable, liang2024gaufregaussiandeformationfields, Wu_2024_CVPR}. Other works regularize the scene flow by imposing a low-rank assumption, representing each Gaussian trajectory by factorizing its motion into a learned motion basis, either local \cite{Lei2024MoScaDG, huang2024sc} or global \cite{som2024, kratimenos2024dynmf, liang2025himor}. To incorporate temporal smoothness, some works represent the Gaussians' motion explicitly by spline or polynomial functions \cite{lin2024gaussian, Li2023SpacetimeGF}. Our work is the first to reconstruct high-speed 3D dynamic scenes from encoded low-speed images by building upon the Gaussian Flow framework \cite{lin2024gaussian}.

\section{Multi-color stroboscopic imaging}
\label{sec:forward}

In this Section, we describe the image formation model of a single camera in our multi-camera system. For simplicity, here we assume all cameras share the same spectral sensitivity function, and discuss color calibration in Sec.~\ref{sec:data_processing_and_calib}.
Consider a scene captured by a color camera and illuminated by a set of \MS{approximately co-located} direct color LEDs.\footnote{\MS{The LEDs must be co-located or sufficiently diffused to avoid shadowing differences between colors.}}
Without loss of generality, suppose that the LED set contains the standard red, green, and blue (RGB) projective color primaries. Furthermore, assume that the scene objects we seek to capture in motion have an approximately uniform spectral reflectance and that the scene background is sufficiently dark such that it reflects a negligible amount of light with respect to the moving object.

Under these assumptions, the camera image $\imgc$, in graylevel units, captured for an exposure duration $T^{\rm exp}$ can be modeled as:
\begin{equation}
    \imgc = \int_{t=0}^{T^{\rm exp}} \ledRGB(t,c) \imgi dt,
    \label{eq:nostrobe}
\end{equation}
where $\pixel$ denotes the image pixel location, $c$ denotes the camera's color channel (\eg, RGB) and $t$ denotes time. The integrand in Eq.~\eqref{eq:nostrobe} is a multiplication of two factors. The first factor, $\ledRGB(t,c), c\in \{{\rm R}, {\rm G},{\rm B}\}$, is the normalized instantaneous color captured by the camera at time $t$, namely:
\red{\begin{equation}
\|\ledRGB(t)\|_2 = 1,
\qquad
\text{where } \ledRGB(t) \in \mathbb{R}^3.
\end{equation}
}
The second factor, $\imgi$, is an instantaneous image intensity having units of $\text{graylevel} \! \cdot\! s^{-1}$, where $s$ is some scaling factor. Note that, under our formulation, the factor $\ledRGB(t,c)$ encapsulates all color-related effects, including the LEDs' illuminant spectra, the object's spectral reflectance, and the camera's spectral response. The second factor, $\imgi$, encapsulates intensity-related effects, such as shading, surface reflectance variations (\ie, texture), scene radiance, and sensor quantum efficiency.

Let $\ledRGBun(t,c)$ denote the unnormalized instantaneous captured object color, namely $\ledRGBun(t,c) \!\!=\!\! s\ledRGB(t,c)$.
Then, $\ledRGBun(t,c)$ can be modeled as
\begin{equation}
    \ledRGBun(t,c) = \alpha(t) \ledR(c) + \beta(t) \ledG(c) + \gamma(t) \ledB(c)
\end{equation}
where $\ledR$,$\ledG$ and $\ledB$ are the individual color primaries corresponding to the red, green and blue LEDs, respectively, and $\alpha(t),\beta(t)$ and $\gamma(t)$ are time dependent LED intensities having a range of $[0,1]$.
These primaries can be calibrated by sampling the RGB camera values of the same small object patch, for three different frames where the object is illuminated by a single LED (\eg, $\alpha\!=\!1$, while $\beta\!=\!\gamma\!=\!0$) keeping the exposure duration constant.

\paragraph{Strobing the scene with a sequence of colors.}
Let $t_n$ denote the strobe start time of the $n$-th strobe, and $T^{\rm strobe}$ denote the strobe duration. During each strobe, the values of $\{ \alpha(t),\beta(t),\gamma(t)\}$ are set to some constants denoted by $\{ \alpha_n,\beta_n,\gamma_n\}$ \red{to yield a sequence of $N$ distinct colors}, while outside the strobing intervals, all values are set to zero (\ie, the LEDs are off). 
Then, Eq.~\eqref{eq:nostrobe} becomes
\begin{align}
        \imgc = \sum_{n=1}^{N}\int_{t=t_n}^{t_n + T^{\rm strobe}}\!\!\!\! \ledRGB(t,c) \imgi dt=~~~~~~~~~~~~\nonumber \\
        \sum_{n=1}^{N} \ledRGB^n(c)\int_{t=t_n}^{t_n + T^{\rm strobe}}\!\!\!\!\imgi dt\approx
        \sum_{n=1}^{N} \ledRGB^n(c)\imgn,
        \label{eq:disframes}
\end{align}
where $\ledRGB^n(c)$ and $\imgn$ are the object color the high-speed intensity frame at strobe $n$. In Eq.~\eqref{eq:disframes} we assume that the strobe duration $T^{\rm strobe}$ is short relative to object motion, having the object approximately static during each strobe.
Later, in Sec.~\ref{sec:HW}, we describe how this assumption allows for setting different intensities for $ \alpha(t)$, $\beta(t)$, and $\gamma(t)$ using only `binary intensity' LEDs. \red{In the last step in Eq.~\eqref{eq:disframes}, we absorbed the constant $T^{\rm strobe}$ in $\imgn$}.


\section{ Volumetric scene reconstruction}
\label{sec:inverse}
\newtheorem{remark}{Remark}

In this section, we describe our novel method for modeling high-speed 3D dynamic scenes by extending an existing dynamic 3DGS method with a modified training process.
For the reader's benefit, we start with a brief description of the dynamic 3DGS method we rely on, followed by a description of the extensions we made to its training process.

\paragraph{Scene representation using Gaussian splatting}
In Gaussian splatting, the 3D scene is represented by a set of Gaussians $G$. Given an arbitrary camera position $\phi\in \mathbb{R}^6$, the set of Gaussians can be used to render an image from view $\phi$ using the rendering function $\mathcal{R}$:
\begin{equation}
    \mathcal{R}(G, \phi) = \imgeilon(c).
    \label{eq:render}
\end{equation}
For convenience, we drop the pixel $\pixel$ notation from $\imgc$ from this point on.
Each individual Gaussian $g{\subset} G$ is defined by a set of parameters (\eg, positions, scales, rotations, colors). Given $M$ input views $\phi_m$ of the scene $\{\imgeilon_m(c,\phi_m)\}_{m=1}^M$ indexed by $m{=}1,2,..M$, the Gaussians in $G$
 are optimized via differentiable gradient-based optimization, such that their `splatting' on the input camera positions $\phi_m$ matches the input images from these views. The 3DGS representation, however, was designed for static scenes. Next, we describe \textit{Gaussian-Flow}, an extension which was designed to handle scene motion \cite{lin2024gaussian}.


\paragraph{Gaussian-Flow} 
In Gaussian-Flow, the Gaussian parameters evolve with time $t$ according to $g(t) {=} g(0) {+} d(t)$, where $g(0)$ is the initial state and $d(t)$ is some chosen deformation function.
Lin \etal, modeled the deformation functions as a sum of a polynomial and a Fourier series, referring to their model as the Dual-Domain Deformation Model (DDDM) \cite{lin2024gaussian}. Since the Gaussians are time-dependent, the optimization now also requires the timestamps of the input frames, \red{$\{t_k\}_{k=1}^K$}.
Let $G(0)$ and $D(t)$ denote the initial state of all the Gaussians and their time-dependent deformations, respectively. Then, the   
Gaussian-Flow objective is:
\red{
\begin{equation}\label{Gaussian-Flow obj}
        \underset{\substack{G(0), D(t)}}{\rm argmin}
        \sum_{k=1}^K\sum_{m=1}^M 
        \left|\mathcal{R}(G(0) {+} D(t_k), \phi_m) {-}  \imgeilon(c,\phi_m,t_k)\right|.
\end{equation}
\vspace{-15pt}
}
\paragraph{Gaussian-Flow with strobing}
\red{
We employ the Gaussian-Flow framework to reconstruct the high-speed scene representation at timestamps $\{t_k{+}t_n\}|_{n=1}^{N}$ for each low-speed timestamp $t_k$. Our optimization does not account for multiple low-speed timestamps and uses only the $M$ frames captured at timestamp $t_k$. Thus, without loss of generality, we set $t_k{=}0$ in the formulation below and drop the frame time from  $\imgeilon(c,\phi_m,t_k)\rightarrow \imgeilon(c,\phi_m)$. The same procedure, described next, is applied to each frame $t_k$ independently. 
}

To model the 3D dynamic scene captured by our system, we make two main modifications to Gaussian-Flow.
First, since the objects we seek to capture are assumed to have uniform spectral reflectance, we change the rendering function $\mathcal{R}(G, \phi)$ to output single-channel intensity images by assigning only a single color channel to each Gaussian.

Secondly, we modify the objective in Eq.~\eqref{Gaussian-Flow obj}. As formalized by Eq.~\eqref{eq:disframes}, each color channel $c$ of an image $\imgeilon(c,\phi_m)$ captured from a camera view $\phi_m$ by our system can be represented as a linear combination of the interframes $\{I^{\text{int}}(\phi_m,t_n)\}_{n=1}^N$.
If $\left(G(0),D(t)\right)$ model the 3D dynamic scene accurately then for the rendered image from view $\phi_m$ at timestep $t_n$, denoted \mbox{$R_n(\phi_m) \equiv \mathcal{R}(G(0) + D(t_n), \phi_m)$}, it follows that $R_n(\phi_m) \approx I^{\textnormal{int}}(\phi_m,t_n)$.
Therefore, by replacing in Eq.~\eqref{eq:disframes} the interframes with the rendered images $\{R_n(\phi_m)\}_{n=1}^N$  we construct an estimator of the captured image 
\begin{equation}\label{eq:approximate by scaled sum}
    \imgeilone(c,\phi_m) \approx
    \sum_{n=1}^N \mathbf{c}_{\textnormal{RGB}}^n(c) \cdot R_n(\phi_m)
\end{equation}
Thus, the training objective minimizes the discrepancy between the estimated and input images
\begin{equation}\label{strobing optimization}
     \underset{\substack{G(0), D(t)}}{\rm argmin}
     \underbrace{\sum_{m=1}^M 
     \sum_{\substack{c\in \{\textnormal{R, G, B}\}}} 
     \left| 
     \imgeilone(c,\phi_m) - \imgeilon(c,\phi_m) 
     \right|}
    _{\mathcal{L}_1}.
\end{equation}

\begin{remark}
    The proposed scheme can be used with various dynamic 3DGS methods.
    We used Gaussian-Flow because of its simple deformation functions, which require training only a few parameters.
    Since the movements in our scenes are rather simple, Gaussian-Flow provides a sufficiently expressive model while keeping the training process efficient.  
\end{remark} 

\begin{remark}\label{remark:single camera}
    The framework above can be applied to the single-camera case as well, yielding high-speed videos from a single encoded frame (see project webpage for a result).
\end{remark} 

\paragraph{Regularization by depth} 
Similar to prior 3DGS methods \citep{chung2024depth}, we additionally incorporate a depth regularization term to promote a geometrically meaningful 3D scene reconstruction. 
Specifically, we penalize the total variation of the inverse depth.  
Let $Z(G,\phi_m,t_n)$ denote the rendered
depth from view $\phi_m$ at timestep $t_n$. The total variation of the inverse depth, $\mathcal{L}_{\text{TV-depth}}$ is 
\begin{align}
    \nonumber &\frac{1}{2 N M} \sum_{m=1}^M \sum_{n=1}^N |\nabla_x\frac{1}{Z(G,\phi_m,t_n)}| + |\nabla_y\frac{1}{Z(G,\phi_m,t_n)}|, 
\end{align}
where $\nabla_x, \nabla_y$ are the gradients computed \red{with} respect to the $x$ and $y$ axis respectively.
Our final loss is therefore
\begin{equation}
      \mathcal{L} = \mathcal{L}_1 + \lambda_{\text{depth}}\mathcal{L}_{\text{TV-depth}},
\end{equation}
where $\mathcal{L}_1$ in defined in \cref{strobing optimization}.

\section{Strobing with color}
\label{sec:strobing}
\figStrob
Given some desired number of interframes $N$, we must select the LED intensities to yield the resulting colors $\ledRGB^n$. The values of $\ledRGB^n$ will affect the reconstruction quality. We tested various methods for selecting the `best' set of colors for $\ledRGB^n$, given $N$. Experimentally, we converged on the color selection in which we sample $\{\alpha_n,\beta_n,\gamma_n\}$ uniformly from a circle in the $\alpha,\beta,\gamma$ space. For the individual LED intensities, this amounts to sampling three sine waves, each separated by a 120-degree phase shift and normalized to the $[0,1]$ range. For a `well-behaved' object reflectance, the resulting circle in $\alpha,\beta,\gamma$ space will approximately map to some ellipse in the camera's RGB space \cite{reinhard2008color}.\footnote{`Well-behaved' in our context refers to an off-white object color having non-negligible reflectance in most of the visible spectrum.}

\section{Data processing and calibrations}
\label{sec:data_processing_and_calib}
\noindent\textbf{Camera color calibration}
In Sec.~\ref{sec:forward}, we assume all cameras share the same spectral sensitivity response. To \red{enforce} this assumption, we calibrate the color response of each camera by imaging a 24-tile Spyder color checker (SCK300) \cite{Datacolor_SpyderCheckrPhoto_SCK300_2025}. Then, we select an arbitrary camera and map all other cameras to the selected camera's color space using a per-camera transformation matrix.

\noindent\textbf{Background subtraction and LED color dictionary calibration}
For each experiment, we capture the static scene before or after the high-speed motion to extract the background images for all frames. We then subtract the background images from the raw frames before applying our strobed Gaussian-Flow model.
The color dictionary $\amat$ is determined directly from the foreground images.

\noindent\textbf{Camera view calibration}
The Gaussian-Flow method requires the camera views as input. We calibrate the cameras using COLMAP by capturing a calibration object without strobing \cite{schoenberger2016sfm}. To initialize the Gaussian-Flow point cloud for dynamic scenes, we run COLMAP on the strobed foreground images. In practice, this initialization suffices.

\section{Imaging prototype}
\label{sec:HW}
\figHW
Our experimental prototype is shown in \cref{fig:HW}. We captured data using 8 global-shutter IDS UI-3240CP cameras, each with a resolution of 1280x1024 \cite{ids}. For each camera, we set the frame rate to 60 FPS (its maximum) and applied a hardware trigger from an Arduino Due \cite{arduinoDue} to synchronize the cameras (discussed further below). The scene is illuminated using a MOBL-300x150-RGBW light \cite{mobl_light}. The light has three-color LED channels (Red, Green, and Blue) that can be individually strobed at high speed.
We built a custom circuit to strobe the light using the Arduino’s digital outputs by generating a PWM signal \MS{for each LED color} from the output pins. A fourth Arduino PWM output signal is used to trigger the cameras.
Our strobing can turn the LEDs on and off, but cannot alter their intensities. We circumvent this limitation and generate different colors by varying the strobe length of each channel, as described below.

As illustrated in \cref{fig:pwm_to_color}, we generate different colors by using various combinations of digital pulse durations. Each combination, integrated over the short pulse duration, yields a distinct color in the captured frame. This type of strobing assumes that the object motion is relatively slow with respect to the strobe duration, which was about 0.083ms on average. Namely, \red{during} each strobe, the scene is approximately static, accumulating the duration of each LED to produce the desired color.
Our results (\eg, Fig. ~\ref{fig:chooper}) show that this assumption holds in our experiments, as no noticeable motion blur artifacts are visible in any experiment.
Each intensity interval lasted approximately 16.7$\mu$s, allowing for six discrete intensity values (from zero to five), yielding $6^3 \!=\! 216$ possible color combinations using the three LEDs. This yields 175 unique, usable colors, excluding scalar multiples of each other and the color no color.

\section{Experimental evaluation}
\label{sec:exp}
\subsection{Real-world experiments}
\figChopper
\figNerf
We demonstrate our approach by capturing rapidly moving objects, such as NERF gun darts, \MS{flying chess pieces}, large particles in motion, and a rotating chopper. Our scenes were selected to be visually similar to the most relevant prior works \cite{chan2023spincam,Sheinin:2021:Deconv} to allow for a qualitative visual comparison.
For each experiment, we capture two videos: one with our strobing system and one with `standard' constant illumination, with all LEDs on. 
In our experiments, the light is set to strobe $N\!=\!10$ colors, which upscales the camera's temporal resolution from 60 to 600 FPS. As shown in the top row of \cref{fig:chooper}, in our experiments, the object motion is too fast for the camera to capture at its maximum speed, resulting in severe motion blur. Our method, however, can recover the high-speed object motion. The project webpage contains additional experiments and volumetric renderings of the captured scenes.

In all experiments, the background pixels are segmented as described in Sec.~\ref{sec:data_processing_and_calib}, and the Gaussian-Flow model is applied only on the foreground images. However, for better visualization, we add the background to some of the reconstructed frames. The background is computed by fitting a static Gaussian splatting model \cite{kerbl20233d} on the background images using the known camera locations. Then, we render the same view from both models (a dynamic foreground and a static background) and combine the frames. Since our background is assumed to be much darker than the dynamic objects, this process yields a reasonable visualization of the full background and foreground scene.

Fig.~\ref{fig:chooper} demonstrates the strength of our method to render novel views at high speeds. Here, eight cameras capture a white sticker on a fast-rotating disk. The figure shows the reconstruction of a \textit{single low-speed frame per camera} into a volumetric high-speed representation of the motion. Rows two and three show the time evolution on two original input views, while rows four and five show novel views. Our method reconstructs the 3D motion with good visual quality (few visible artifacts). We repeated this experiment with a yellow sticker, yielding similar results and showing that the method can handle non-white albedo (see webpage).

In \cref{fig:nerf}, we show experiments representative of motion capture \cite{Sheinin:2021:Deconv,chan2023spincam}. In \cref{fig:nerf}(Top), we shoot a Nerf dart at a board (located on the left of the frame) and capture its motions as it impacts and bounces back. The result shows the dart bouncing off the left wall of the frame. 
Here we show rendered frames from the dynamic Gaussian-Flow model with and without the added background, for reference. Again, we can render novel and existing views during the high-speed dart motion.
In \cref{fig:nerf}(Bottom), we toss several \MS{chess pieces} and capture their motion. Here, we demonstrate the capacity to reconstruct the motion of several larger objects that can overlap.


\subsection{Simulation-based performance evaluation}
\figSimulation
We analyzed the performance and limitations of our method using simulated scenes. 
Our aim here is to gather insights and trends about the method's performance, not to characterize the performance of any one particular camera. \MS{All plots display the Mean Absolute Error (MAE) across all interframes, using a novel camera view as the ground truth.}

\vspace{-12pt}\paragraph{Increasing the number of interframes:}
As shown in Fig.~\ref{fig:simulation}(a), increasing the number of interframes $N$ decreases the cosine distance between the resulting $\alpha_n,\beta_n,\gamma_n$ vectors in $\alpha\beta\gamma$ space and the resulting camera RGB space, making the recovery more susceptible to noise.
In addition, as the number of interframes in a single frame increases, their overlap also increases, making decoupling the interframes challenging.  
The plot shows performance degradation near $N{=}28$, leading to artifacts. 

\vspace{-12pt}\paragraph{Effect of ambient light:}
Our method is sensitive to ambient light, reducing the $\ledRGB^n$ color contrast and potentially causing saturation during long exposures. In \cref{fig:simulation}(b) we simulate this effect by adding ambient signal to the input, $I^c + c_{\rm amb} I^{\rm DC}$, where $I^{\rm DC}$ is the scene under constant white light and $c_{\rm amb}$ controls its strength. We assume ambient light affects only the foreground, since background pixels can be removed via subtraction. Results show a roughly linear increase in reconstruction error with ambient-light level.

\vspace{-12pt}\paragraph{Effect of object albedo:}
\cref{fig:simulation}(c) shows how object reflectance affects reconstruction quality. We simulate an albedo that transitions from pure white to pure red using a blending parameter $\alpha_{\rm albedo}$. The method performs well for broad-spectrum (off-white) albedos but degrades as the object becomes less reflective in one or more spectral bands.

\vspace{-12pt}\paragraph{Number of cameras:}
In \cref{fig:simulation}(d), we analyze the reconstruction quality as a function of the number of cameras $M$. 
The plot indicates that our method can synthesize unseen views reliably with as few as six cameras.
\red{
\vspace{-12pt}\paragraph{Sensitivity to motion complexity:}
We evaluated recovery performance as interframe motion becomes less smooth. 
To this end, we simulated a scene in which motion starts as a simple line and progressively becomes more erratic by randomly sampling the interframe y-coordinate as \mbox{$y \sim \mathcal{U}(-n, n)$}. 
Fig.~\ref{fig:simulation}(e) shows that for high-variance motion, the method may fail by “dropping” several interframes. 
We believe this is partially due to vanishing Gaussian gradients under large displacements \cite{charatan23pixelsplat}.
}

\section{Discussion and limitations}
\label{sec:disc}
\paragraph{Specialized scenes and beyond.}
\MS{
Our current method assumes uniform albedo objects with spatially varying shading. Future extensions could relax this constraint by explicitly modeling the object surface's spatially varying spectral response.\footnote{Spectral response here accounts for both the object's spectral reflectance and the camera's spectral sensitivity functions.} 
For example, as the object moves during the strobing, the same surface patch may appear at different image locations under multiple known strobes; aggregating these observations across interframes could enable joint recovery of motion and surface appearance.
Moreover, combining white-light strobes with cameras operating at different exposure timings could produce strobed frames that enable recovery of scenes with spatially varying albedo. Adding data-driven priors that regularize object shape, texture distributions, and physically plausible dynamics could also help extend the method beyond specialized scenes.
}

\vspace{-12pt}\paragraph{Encoding with color: strengths and weaknesses.}
As in some prior works, we encode high-speed temporal information using color \cite{chan2023spincam,Sheinin:2021:Deconv,stroboscopic_microscopy_3_hue}. However, in contrast to earlier works, our method requires no mechanically moving parts \cite{chan2023spincam}, specialized optics \cite{Sheinin:2021:Deconv}, and involves multi-view optimization-based volumetric scene reconstruction capable of temporal up-scaling by factors of more than three, across multiple views \cite{stroboscopic_microscopy_3_hue}.
While unmixing the strobed input frames is an ill-posed problem, the multi-view solver implicitly leverages epipolar geometry constraints, thereby yielding more robust solutions.

Nevertheless, encoding with color has disadvantages. Firstly, the uniform albedo assumption limits applicability to specialized scenes. The light-squared-distance fall-off may degrade SNR for distant objects (common to all active methods), especially when strobing with color. Nevertheless, integrating multiple frames into a single exposure reduces read noise per frame \cite{schechner2007multiplexing}; in our method, the light source is decoupled from the cameras and can be placed close to the scene.\footnote{In fact, multiple light sources can be used at once as long as they are synchronized temporally.} 
Currently, our experiments require a black background, but our simulations show promise for handling non-dark backgrounds too (see project webpage).
\MS{
Lastly, unlike the standard Gaussian-Flow objective, which uses the object's color via the spherical harmonics loss term, our objective operates on \textit{monochrome} interframes, making it less robust to recovering colorful objects.
Nevertheless, this limitation diminishes when objects are predominantly single-colored, as is our premise.
}

\vspace{-12pt}\paragraph{Interframe temporal resolution.}
The temporal upsampling we achieve depends on our capacity to decompose the mixtures of frame colors. The more interframes we have, the less distinct the resulting colors become in the camera's RGB space, increasing the susceptibility to noise and reconstruction artifacts. This is exacerbated by non-white object albedos, which can reduce the separation in the RGB space. 
However, since strobing LEDs at high speeds (\ie, hundreds of KHz) is much easier than increasing a camera's frame rate, our method could be used \MS{to upscale the speed of faster cameras too, using the same strobe number (\eg, 60 FPS to 600 FPS, 600 FPS to 6000 FPS).}

\red{
\vspace{-12pt}\paragraph{Cameras and light synchronization.}
All cameras and LEDs are driven by one Arduino. 
To continuously strobe $N$ interframes during $T^{\rm exp}$, the strobe spacings yield a margin of $T^{\rm marg} {\equiv} T^{\rm exp} / (2N)$ at both exposure boundaries (see Fig.~\ref{fig:pwm_to_color}). Thus, each camera's trigger can drift up to $T^{\rm marg}$ without affecting the captured frame. For us, $T^{\rm marg} {=} 0.83$ms, which is $\sim$27$\times$ larger than the typical $30\mu$s hardware trigger jitter reported for our cameras \cite{liu2023embeddedGNSS}. This margin eliminates frame-strobe mis-synchronization risk.}


\section{Conclusion}
\label{sec:conclusion}
We presented a novel framework for volumetric high-speed scene reconstruction using conventional low-speed cameras. 
By encoding temporal information directly into the scene via high-frequency color strobes, our method multiplexes several high-speed interframes into each captured image without requiring specialized sensors or complex optical components. We also developed a novel approach that uses a dynamic Gaussian-flow representation. Our approach enables the recovery of high-speed 3D motion from multi-view observations, effectively bridging compressive video techniques with modern volumetric rendering. 

\paragraph{Acknowledgments:}
We thank Tali Dekel for helpful discussions.
This work was supported by the Shimon and Golde Picker – Weizmann Annual Research Grant.

\clearpage
{
    \small
    \bibliographystyle{ieeenat_fullname}
    \bibliography{main}
}

\end{document}